%% file: main.tex
\documentclass[conference]{IEEEtran}
\IEEEoverridecommandlockouts
\usepackage{cite}
\usepackage{amsmath,amssymb,amsfonts}
\usepackage{algorithmic}
\usepackage{algorithm}
\usepackage{graphicx}
\usepackage{tabularx}
\usepackage{textcomp}
\usepackage{xcolor}
\usepackage{stfloats}
\usepackage{booktabs} 
\usepackage{url}
\usepackage{verbatim}
\usepackage{nccmath}
\usepackage{mathtools}
\usepackage{array}
\usepackage[margin=1in]{geometry}

\newcolumntype{C}[1]{>{\centering\arraybackslash}p{#1\linewidth}}
\newcolumntype{L}[1]{>{\raggedright\arraybackslash}p{#1\linewidth}}
\newcommand{\bb}[1]{\boldsymbol{#1}}
\DeclareMathOperator*{\argmax}{arg\,max}
\DeclareMathOperator*{\argmin}{arg\,min}
\def\BibTeX{{\rm B\kern-.05em{\sc i\kern-.025em b}\kern-.08em
    T\kern-.1667em\lower.7ex\hbox{E}\kern-.125emX}}
\begin{document}

\title{Learning Human Reaching Optimality Principles from Minimal Observation Inverse Reinforcement Learning}

\author{Sarmad Mehrdad$^{1}$, Maxime Sabbah$^{2}$, Vincent Bonnet$^{2,3}$, Ludovic Righetti$^{1,4}$

\thanks{${^1}$  Machines in Motion Laboratory, New York University, USA}
\thanks{${^2}$ LAAS-CNRS, Université Paul Sabatier, CNRS, Toulouse, France.}
\thanks{${^3}$Image and Pervasive Access Laboratory (IPAL), CNRS-UMI, 2955, Singapore.}
\thanks{${^4}$ Artificial and Natural Intelligence Toulouse Institute (ANITI), Toulouse}
}

\maketitle

\begin{abstract}
\input{abstract/abstract}
\end{abstract}

\section{Introduction}
\input{introduction/introduction}

\section{Methods}
\input{methods/methods}

\section{Results}

\input{results/results}

\section{Conclusion} 
\input{conclusion/conclusion}

\section{Acknowledgments}
The authors would like to deeply thank Dr. Bastien Berret for kindly sharing his experimental data and thorough discussions on the topic. 

\bibliographystyle{IEEEtran}
\bibliography{ref}
\end{document}

%% file: abstract/abstract.tex
This paper investigates the application of Minimal Observation Inverse Reinforcement Learning (MO‑IRL) to model and predict human arm‑reaching movements with time‑varying cost weights. Using a planar two‑link biomechanical model and high‑resolution motion‑capture data from subjects performing a pointing task, we segment each trajectory into multiple phases and learn phase‑specific combinations of seven candidate cost functions. MO‑IRL iteratively refines cost weights by scaling observed and generated trajectories in the maximum entropy IRL formulation, greatly reducing the number of required demonstrations and convergence time compared to classical IRL approaches. Training on ten trials per posture yields average joint‑angle Root Mean Squared Errors (RMSE) of 6.4 deg and 5.6 deg for six‑ and eight‑segment weight divisions, respectively, versus 10.4 deg using a single static weight. Cross‑validation on remaining trials and, for the first time, inter‑subject validation on an unseen subject’s 20 trials, demonstrates comparable predictive accuracy, around 8 deg RMSE, indicating robust generalization. Learned weights emphasize joint acceleration minimization during movement onset and termination, aligning with smoothness principles observed in biological motion. These results suggest that MO‑IRL can efficiently uncover dynamic, subject‑independent cost structures underlying human motor control, with potential applications for humanoid robots.

%% file: introduction/introduction.tex
Understanding the optimal principles underlying simple motions like human arm reaching is crucial for progress in both neuroscience and robotics. In neuroscience, these principles shed light on how the central nervous system plans and executes goal-directed movements under constraints such as muscle redundancy, sensory noise, and biomechanical limitations. Capturing these strategies helps elucidate motor control mechanisms and supports clinical rehabilitation by identifying deviations from optimality in pathological movements.
In robotics and human–robot interaction, modeling human reaching as an optimal control problem facilitates the design of bio-inspired controllers and predictive algorithms. This is especially valuable for humanoids, assistive robotics and prosthetics, where human-likeness and intent prediction are critical.
Biological motion exhibits invariant properties despite the wide range of available motor strategies. Voluntary movements tend to follow consistent patterns, suggesting that the nervous system resolves motor redundancy by adhering to specific organizational principles, the so-called optimal weights.
However, the precise link between cost functions and variables encoded by the central nervous system remains unclear. 
Moreover, the idea of a single universal cost function may be unrealistic. 
The central nervous system might flexibly adjust cost weightings based on task demands \cite{cao2024}, or even during the same task. For example, individuals may reduce velocity at the end of a reach to aim more accurately, while seeking overall speed. This suggests a balance between objective (task-related) and subjective (body-related) costs, which current models often fail to capture \cite{zelik2012}.

A widely used framework for exploring the principles underlying motor control is optimal control theory, which makes the hypothesis that biological movements arise from the minimization of specific cost or loss functions. Numerous models based on this theory have been proposed \cite{berret2011, sylla2014}, many of which claim to replicate experimental data with reasonable accuracy. However, these models often rely on a single cost function per task, which may not adequately capture the complexity and variability of human motion. As a result, relatively high Root Mean Square Errors (RMSEs) are frequently observed between the predicted and measured trajectories. For instance, Sylla et al. \cite{sylla2014} reported an average RMSE of $7$deg, with some angles exhibiting errors superior to $15$deg,  even for simple reaching movements. Moreover, many studies unfortunately do not report any quantitative comparison or use tailored metrics \cite{berret2011} between their model predictions and experimental data. This raises questions about the relevance and predictive power of such models, especially when considering the sensitivity of their outcomes to variations in the chosen cost function components. 

Unfortunately, because of the current limitation of  Inverse Optimal Control (IOC) and Inverse Reinforcement Learning (IRL) methods used to retrieve optimal cost function weights from human optimal motion, a single set of parameters for a given task is generally used \cite{Lin2021}.
Indeed, adding time-varying weights leads to a significant increase in the number of parameters to be identified which these algorithms struggle to handle. In this paper, we leverage a new
efficient IRL algorithm to instead study how time-varying weights lead to a more nuanced and accurate description of the movement.


IOC provides a model-based framework for inferring  cost function weights that best explain observed human motion, assuming that the motion is optimal for some performance criterion \cite{Lin2021}. Despite its conceptual appeal, practical applications of IOC face significant challenges. The standard bi-level formulation, in which cost weights are optimized through repeated solutions of a nested optimal control problem, is computationally expensive, often requiring several days of computation. Additionally, this approach is prone to convergence to local minima, particularly in high-dimensional problems.
To address these issues, alternative formulations based on the residuals of the Karush-Kuhn-Tucker (KKT) conditions have been proposed. These methods aim to eliminate the need for repeated trajectory optimization. However, they remain highly sensitive to measurement noise and modeling errors commonly observed in human motion data \cite{colombel2022, bevcanovic2022}. More recently, promising hybrid approaches that combine elements of the bi-level and residual-based formulations have been introduced \cite{becanovic2024reliability}, although these have only been validated in simulation.

In contrast to IOC, IRL adopts a probabilistic framework to infer the underlying cost function. This approach relaxes the number of bi-level-like iterations and is especially appealing for tasks involving uncertainty and variability, such as those performed by humans. IRL defines a probability distribution over all demonstrations and seeks to identify the cost function that maximizes the likelihood of the optimal trajectories. Given this definition, IRL ideally requires all the possible trajectories for the utmost optimal cost function derivation, which is impossible. Hence, IRL's performance is heavily hinged on the trajectory space approximation accuracy obtained from a finite set of observations. There have been several efforts to circumvent this shortcoming by approximating the trajectory set through more intelligent sampling around the optimal trajectory \cite{kalakrishnan2011, kalakrishnan2013} and trajectory set augmentation \cite{finn2016}.

However, the sampled trajectories often lie close to the observed ones and may not sufficiently explore the broader trajectory space. Furthermore, in order to improve cost function estimation, IRL must consider all sampled and iteratively generated trajectories in the probability maximization process. This requirement substantially increases the computational cost of IRL. To address these shortcomings, we turn to the newly proposed Minimal Observation Inverse Reinforcement Learning (MO-IRL) \cite{mehrdad2025}. MO-IRL takes an iterative approach for cost function estimation, by approximating the trajectory space through scaling the effectiveness of each observed trajectory depending on its current estimate of optimality. With this added feature, even with a small observation set, MO-IRL empirically provides better iterates that lead to an improved estimation of the weights. This leads to iterative cost function learning with minimal information about the trajectory space, resulting in considerably faster convergence. To our knowledge, MO-IRL was designed and tested only with robotics tasks and fixed weights.

In this paper, we extend MO-IRL to learn tasks requiring time-varying cost weights and investigate its use in predicting accurate human joint trajectories by learning simultaneously from positions and velocities. The proposed approach is validated with a subset of reference human data from the human motor control community \cite{berret2011}. In particular, we show that the method can learn task weights leading to accurate movement reproduction that also generalize across movements.


%% file: methods/methods.tex
\subsection{Experimental protocol and mechanical model}
\begin{figure}
    \centering
    \includegraphics[width=0.9\linewidth]{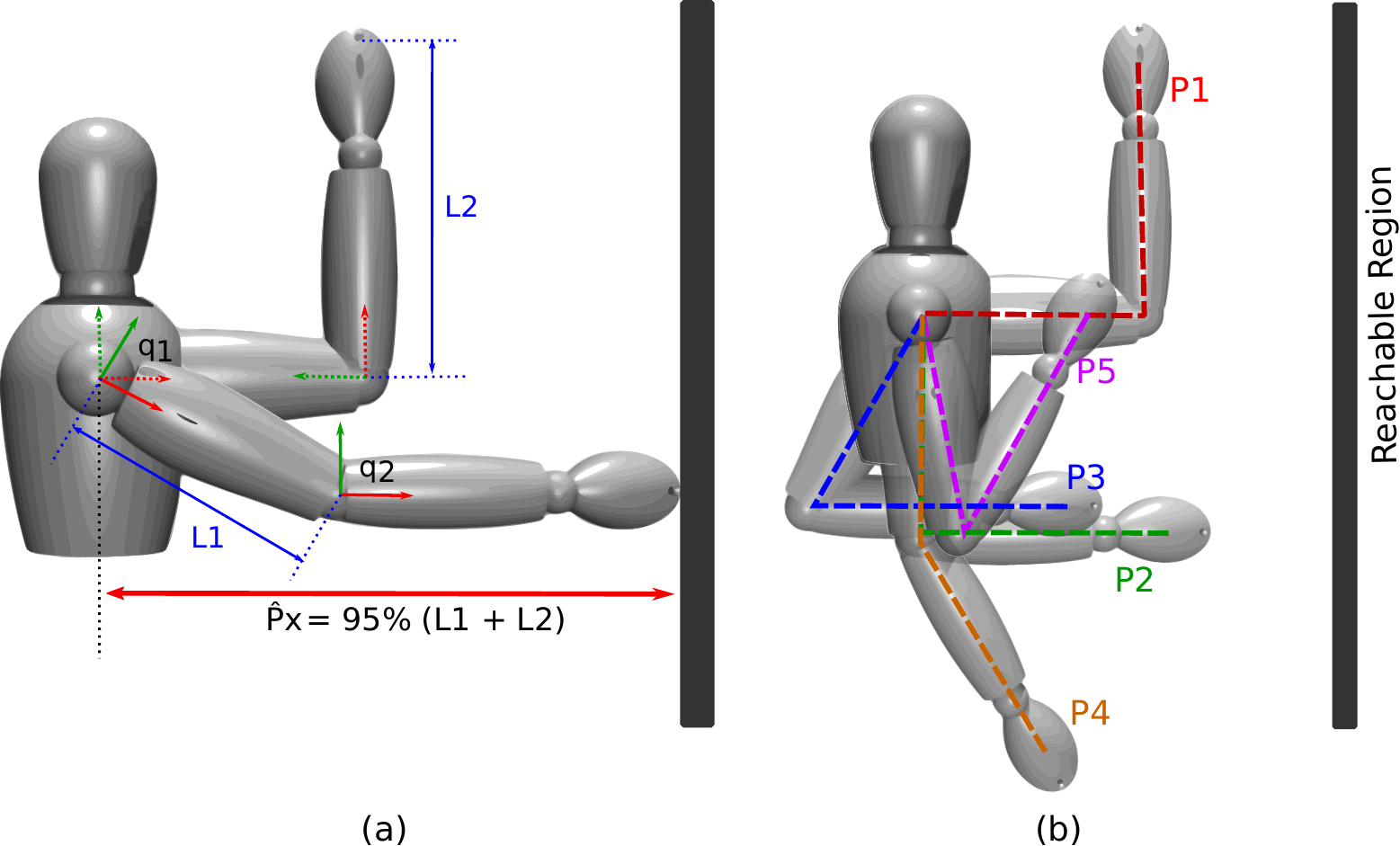}
    \caption{(a) Biomechanical model definition, showing the beginning and the end of the pointing task. (b) Five different initial postures for the pointing task \cite{berret2011}.}
    \label{fig:biomec_model}
\end{figure}

The human data used in this study were kindly provided by Berret al. \cite{berret2011}. These data were used in numerous other studies since their publication and are considered a reference. They consist of motion capture 3D marker positions, including markers located at the shoulder, elbow, and wrist level. Data from twenty right-handed naive subjects were provided. For this preliminary study, the data of only two subjects were selected. Subjects first had to sign the approved local ethical committee ASL-3 ("Azienda Sanitaria Locale", local health unit), Genoa, Italy. 
Then, they performed the pointing task as depicted in Fig. \ref{fig:biomec_model}.a. While seated, participants were instructed to perform a series of pointing movements toward a vertical target bar positioned in front of the participant. Only shoulder and elbow flexion/extension were permitted during the task, as wrist movement was restricted.  The shoulder-to-bar horizontal distance was set to 95\% of the participant’s total arm length ($L = L_1 + L_2$, with $L_1$ and $L_2$ representing upper arm and forearm lengths, respectively; Fig. \ref{fig:biomec_model}.a). Five initial arm postures, labeled P1 through P5, were defined using reference points positioned in a vertical plane located approximately 10 cm lateral to the right shoulder. These five postures corresponded to specific predefined angular configurations of the arm and are shown in Fig. \ref{fig:biomec_model}.b. Each individual performed 20 trials for each posture.


A planar biomechanical model, illustrated in Fig.\ref{fig:biomec_model}.a, was developed to represent flexion/extension movements at the shoulder ($q_1$) and elbow ($q_2$) joints. The model’s base frame was located at the shoulder joint, and the relative positions of successive joints in their parent frames were computed using segment lengths $L_1$ and $L_2$, estimated from marker data. Inertial parameters were calculated using anthropometric tables \cite{Dumas2007}.

\subsection{Optimal control problem}
In the context of pointing or reaching movements, Berret et al. \cite{berret2011} proposed a set of $N_\Phi = 7$ candidate cost functions, as detailed in Table \ref{Table:cost_functions}. Although the task may appear elementary, we posit that individuals do not adhere to a single cost function throughout the entire movement. For instance, it is intuitive to expect a deceleration near the endpoint to ensure accurate and controlled pointing. To account for such time-varying motor strategies, each recorded trajectory of duration $T$ was segmented into $N_w$ equal time windows, each comprising $N_s$ samples. This segmentation was chosen based on consistent inflection points observed in the majority of trajectories.
To model the temporal evolution of movement strategies, we introduced a weight matrix $\bb{\omega} \in \bb{R}^{N_\Phi \times N_w}$, which allows distinct cost function contributions across different movement phases. The full trajectory $\bb{x} \in \bb{R}^{N_s\times N_w}$ was defined as the concatenation of state vectors $\bb{x}_s = (\bb{q}_s, \dot{\bb{q}}_s)$ for each section $s \in \{1,\cdots,N_w\}$. Accordingly, $\bb{u}_s$ is defined as the control torque input to the human model joints for each section. The associated Direct Optimal Control (DOC) problem was then formulated to reconstruct the observed human motion over this multi-phase framework.

\begin{equation}
\begin{aligned}
\bb{x}^* = \ & \argmin_{\bb{u}} \quad \sum_{s=1}^{N_w}\sum_{j=1}^{N_\Phi} \omega_{s,j} \Phi_{j}(\mathbf{x}_s, \bb{u}_s) \\
\text{s.t.} & \quad \bb{x}(t+1) = f(\bb{x}(t), \bb{u}(t)), \\
            & \quad P_X(T) = \hat{P}_{X}, \\
            & \quad \bb{q}^-\leq \bb{q} \leq \bb{q}^+\\
            & \quad \bb{q}(0)=\bb{q_0}\\
            & \quad ||\bb{\dot{q}}||\leq\bb{\dot{q}}^+\\
            & \quad \bb{\dot{q}}(0)=\bb{\dot{q}}(T)=0
\end{aligned}
\label{eq:doc}
\end{equation}

\noindent where $\bb{x}(t)$ and $\bb{u}(t)$ are the state and control at discrete time $t$; $f$ is the Euler time-discretized dynamics; $P(t)$ is the position of the hand  obtained from forward kinematics, and $\hat{P}_X$ is the goal position on the horizontal axis the subject aims to reach as shown Fig. \ref{fig:biomec_model}.a; $ \bb{q_0}$ is the initial human joint configuration; $\bb{q}^-$, $\bb{q}^+$ are the lower and upper joint boundaries respectively; $\bb{\dot{q}}^+$ is the maximal joint velocity. 

We used Pinocchio \cite{pin} for modeling the human body, together with the Croccoddyl framework \cite{crocoddyl} and the MiM\_Solver nonlinear CSQP solver \cite{sqp} to define and solve the constrained DOC in Eq.\eqref{eq:doc}. We used MuJoCo \cite{mujoco} for the model simulation.


\begin{table}[!t]
\centering
\fontsize{7}{11}\selectfont
\caption{(Discretized) Biomechanical cost functions \cite{berret2011}}
\begin{tabular}{cccc} \toprule
\textbf{Label} &\textbf{Name} & \textbf{Equation} & \textbf{Reference} \\ \midrule
$\Phi_1$    & Cartesian velocity     &  $ \sum_{t=0}^{T} \dot{P}(t)^T\dot{P}(t) \,dt $ & \cite{flash1985}     \\
$\Phi_2$    & Energy     &  $ \sum_{t=0}^{T} \left|\dot{\mathbf{q}}(t)^T \mathbf{u}(t)\right| \,dt $ & \cite{nishii2002, berret2008}     \\
$\Phi_3$    & Geodesic     &  $ \sum_{t=0}^{T} \dot{\mathbf{q}}(t)^TM \dot{\mathbf{q}}(t) \,dt $ & \cite{biess2007}     \\
$\Phi_4$    & Joint acceleration     &  $ \sum_{t=0}^{T} \ddot{\mathbf{q}}(t)^T\ddot{\mathbf{q}}(t) \,dt $ & \cite{ben2008}     \\
$\Phi_5$    & Joint torque change     &  $ \sum_{t=0}^{T} \dot{\bb{\tau}}(t)^T\dot{\bb{\tau}}(t) \,dt $ & \cite{uno1989,nakano1999}    \\
$\Phi_6$    & Joint velocity     &  $ \sum_{t=0}^{T} \dot{\mathbf{q}}(t)^T\dot{\mathbf{q}}(t)  \,dt $ & \cite{atkeson1985}     \\
$\Phi_7$    & Joint torque     & $ \sum_{t=0}^{T} \bb{\tau}(t)^T\bb{\tau}(t)  \,dt $ & \cite{nelson1983}     \\ \bottomrule
\end{tabular}
\label{Table:cost_functions}
\end{table} 

\subsection{Minimum Observation Inverse Reinforcement Learning (MO-IRL)}

As mentioned before, we aim to solve an inverse optimal control problem to derive the optimal cost function as a linear combination of explicit features. 
For this purpose, we use an augmented MO-IRL\cite{mehrdad2025} framework to accommodate learning for several time windows with different weights. As commonly used in IRL algorithms, MO-IRL aims to maximize the probability of the optimal demonstrations (i.e. the human data)

\begin{align}
    \label{maxent}
    &\bb{\omega}^* = \argmax_{\bb{\omega}} P(\bb{x}^*|\bb{\omega} , \Bar{\bb{x}}) \\
    \text{where} \hspace{5mm} &P(\bb{x}^*|\bb{\omega} , \Bar{\bb{x}}) = \frac{e^{-\bb{\omega}^{T}\bb{\Phi}^*}}{\sum_{i=1}^{K} e^{-\bb{\omega}^{T}\bb{\Phi}_i}} \nonumber \\
    &\bb{\omega} \geq 0 \nonumber
\end{align}

\noindent
in which $\Bar{\bb{x}}$ is the set of $K$ observed trajectories. For brevity, we write $\bb{\omega}$ the concatenated weight vector for the cost features and $\bb{\Phi}$ the concatenated feature vector. In the following, $\bb{\Phi}(\bb{x}_i, \bb{u}_i)$ which is the feature costs for the $i$\textsuperscript{th} trajectory is henceforth referred to as $\bb{\Phi}_i$. 
As observed in \cite{mehrdad2025}, all the sub-optimal trajectories are being incorporated and contribute equally in the denominator, irrespective of how close they are to optimality. This lack of distinction can create numerical issues for the optimizer. Therefore, it is important to scale them to emphasize their effectiveness so the optimizer will have better information about what the approximated trajectory set represents in terms of optimality. 

MO-IRL solves Eq.  \eqref{maxent} by iteratively improving $\bb{\omega}$ rather than optimizing it in one shot. Considering an update of cost weights at each iteration $n+1$ in the form $\bb{\omega}_{n+1} = \bb{\omega}_n + \Delta\bb{\omega}_n$, the original probability distribution can be rewritten to instead find the best $\Delta\bb{\omega}_n$:

\begin{align}
    \label{'delta_w'}
    \Delta\bb{\omega}_n^* &= \argmin_{\Delta\bb{\omega}_n} -\log \frac{1}{1 + \sum_{\bb{x}_i \in \Bar{\bb{x}}} \gamma_i e^{-\Delta\bb{w}_t^T(\bb{\Phi}_i - \bb{\Phi}^*)}} \nonumber\\
    \text{s.t.}  \hspace{3mm} &\Delta\bb{\omega}_n > -\bb{\omega}_n \\
    &\gamma_i = e^{-\bb{\omega}_n^T(\bb{\Phi}_i - \bb{\Phi}^*)} \nonumber
\end{align}

\noindent
In this case, sampled trajectories are automatically scaled depending on their cost in the previous iteration. MO-IRL solves Eq. \eqref{'delta_w'} and then seeks to find an update of the form $\omega_{t+1} = \omega_t + \alpha\Delta\omega$ where $\alpha$ is selected using a merit function (similar to a line search procedure). Starting with $\alpha = 1$, the algorithm checks if the resulting trajectory is closer to the optimal demonstration by evaluating the merit function. If the merit function value has not been decreased with the added change to the weight, MO-IRL scales down $\alpha$ by factor of $0.25$, and tries again for a maximum of 10 trials. If by the 10\textsuperscript{th} trial there was no improvement, the algorithm stops. Otherwise, the accepted trajectory is added to the observed trajectory set $\Bar{\bb{x}}$, and MO-IRL moves on to the next iteration.

In the literature, algorithms to learn human motion trajectories usually minimize the gap between the estimated and the optimal trajectories in joint space without considering velocities. In this study, however, we propose to minimize the gap in both joint position ($\bb{q}$) and joint velocity ($\bb{\dot{q}}$) concurrently. Therefore, we evaluate the estimation improvement based on the full state vector $\bb{x} = [q_1, q_2, \dot{q}_1, \dot{q}_2]$. We define the merit function as $m(\bb{x}) = \frac{1}{T}||\bb{x}^* - \bb{x}||^2_2$.
For this study, we discard all previously generated non-optimal trajectories from $\Bar{\bb{x}}$ and use only the most recent generated non-optimal trajectory as the trajectory set, i.e., $\Bar{\bb{x}} = \{\bb{x}_t\}$ as we empirically noticed that it leads to faster convergence. Initial weights are set to small uniform value ($\bb{\omega} = 0.05$).

We extend MO-IRL for learning multiple weight sections from multiple demonstrations:
\begin{align}
    \label{'moirl'}
    \small
    &\Delta\omega_t^* =  \argmin_{\Delta\omega_t} \nonumber \\
    &\sum_{d = 1}^D \bigg( -\log \frac{1}{1 + \sum_{\bb{x} \in \Bar{\bb{x}}} \gamma_i e^{-C(\bb{x}_i, \Delta\bb{\omega}_t)}}\bigg) + \frac{\beta}{2}||\Delta\omega_t||_2^2 \nonumber \\
    &\text{s.t.}  \hspace{3mm} \Delta\bb{\omega}_t > -\bb{\omega}_t  \\
    &\hspace{7mm} \gamma_i = e^{-C(\bb{x}_i, \omega_t)} \nonumber \\
    &\hspace{7mm} C(\bb{x}_i, \Delta\omega_t) = \sum_{s = 1}^{N_\omega} \Delta\bb{\omega}_{st}^T(\Phi_{si} - \Phi_{sd}^*) \nonumber
    \normalsize
\end{align}
%
where $D$ is the number of provided optimal demonstrations, and $N_\omega$ is the number of weight sections as mentioned before. We also use a small $L2$ regularizer ($\beta = 10^{-10}$) for the optimization to prevent high changes in weights and overfitting. When learning from multiple demonstrations, the merit function for  step acceptance is changed to $m(\bb{x}) = \frac{1}{D}\sum_{d = 1}^D(\frac{1}{N}||\bb{x}_d^* - \bb{x}||^2_2)$. 

\subsection{Learning and Cross-Validations}

This section describes the learning and cross-validation processes of the proposed method. The human trial dataset consists of 20 trials of the wall-reaching task for five initial postures for 2 human subjects. In addition to accuracy, we aim to test the generalizability of the learned weights, i.e. can one subject's trials be informative enough to predict other trials. We test our algorithm by comparing the joint value RMSE of the generated trajectory by the DOC using the learned weights against optimal trajectories executed by the human subject. 


Cost weights are learned for each posture using 10 randomly selected trials from one subject and are cross-validated on the remaining 10 trials. The quality of the learned weights is evaluated by the average RMSE between the trials and the trajectories generated by the DOC using the learned weights. 
For further assessment of the MO-IRL's efficacy, we also perform an Inter-Subject cross-validation (ISCV) to see how the learned weights predict another subject's motion for the same posture. 
This helps evaluate whether the learned weights can generalize to another individual. 


%% file: results/results.tex
We evaluate the results for the two subjects, one for training and cross-validation, and another for ISCV, from the dataset provided by Berret et al. \cite{berret2011}. 
We evaluate our framework for each initial posture using 1, 6, and 8 weight sections. Note that for each posture, weights are identified separately as different strategies per posture were suggested by Berret et al. \cite{berret2011}. 

\begin{figure*}[t]
    \centering
    \includegraphics[width=0.89\linewidth]{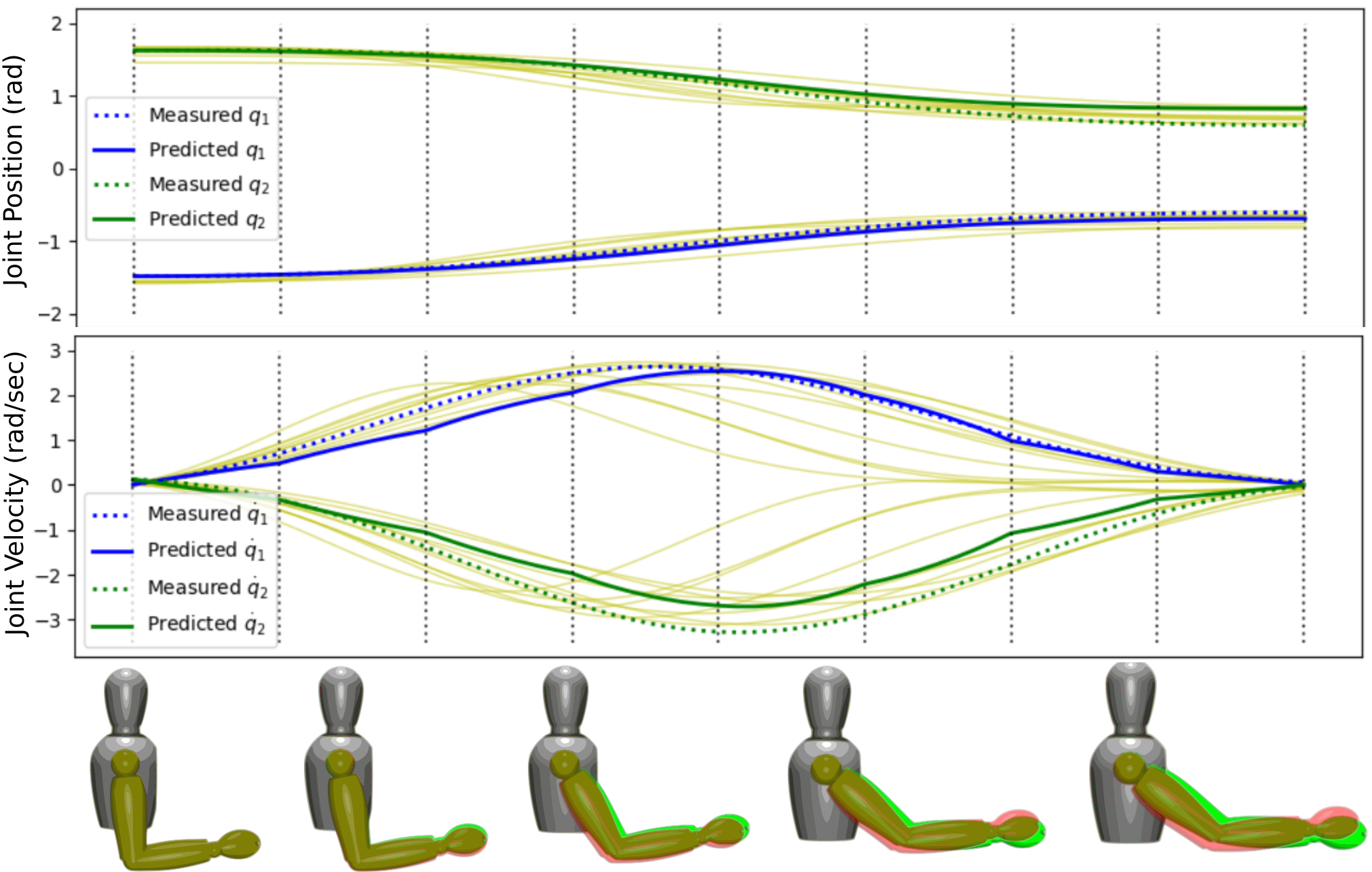}
    \caption{Illustration of MO-IRL prediction against the actual human task execution. The cost weights are divided into 8 sections. The training data for both joint positions and velocities are shown in yellow. The dotted lines are the real trajectory performed by the human, and the solid lines are the MO-IRL predictions.}
    \label{fig:prediction}
\end{figure*}

Fig.~\ref{fig:prediction} illustrates a comparison between the measured and estimated joint trajectories for posture 2, after learning the weights in 8 sections. The figure reveals that while the human demonstrations are generally similar, they exhibit noticeable variability, particularly in joint velocities, which vary more than joint positions. This observation supports our claim regarding multi-modality in IRL. Such variability is beneficial for MO-IRL, as it prevents the algorithm from fitting a cost function that reproduces a single trajectory too closely. Instead, it favors trajectories that lie within the general vicinity of the demonstrated motions. This mitigates overfitting, especially in the case of joint positions, which tend to be more consistent across trials. Furthermore, subtle changes that could be expected from segmenting the weights are visible as shown in the joint velocity profiles.

\begin{figure*}[!ht]
    \centering
    \includegraphics[width=0.9\linewidth]{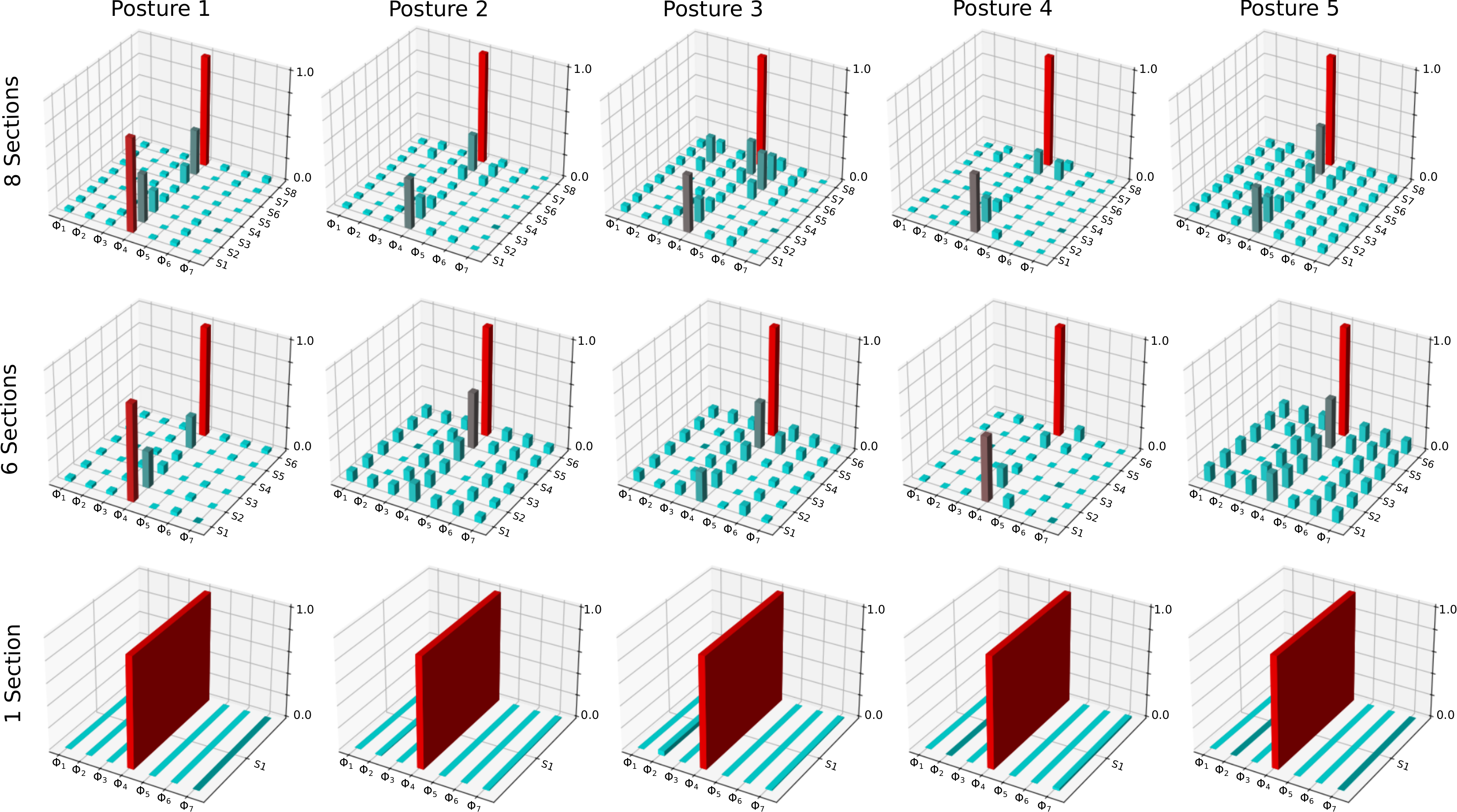}
    \caption{Normalized weights learned by MO-IRL for each posture given 1, 6, and 8 sections.}
    \label{fig:weights}
\end{figure*}

\begin{figure}
    \centering
    \includegraphics[width=\columnwidth]{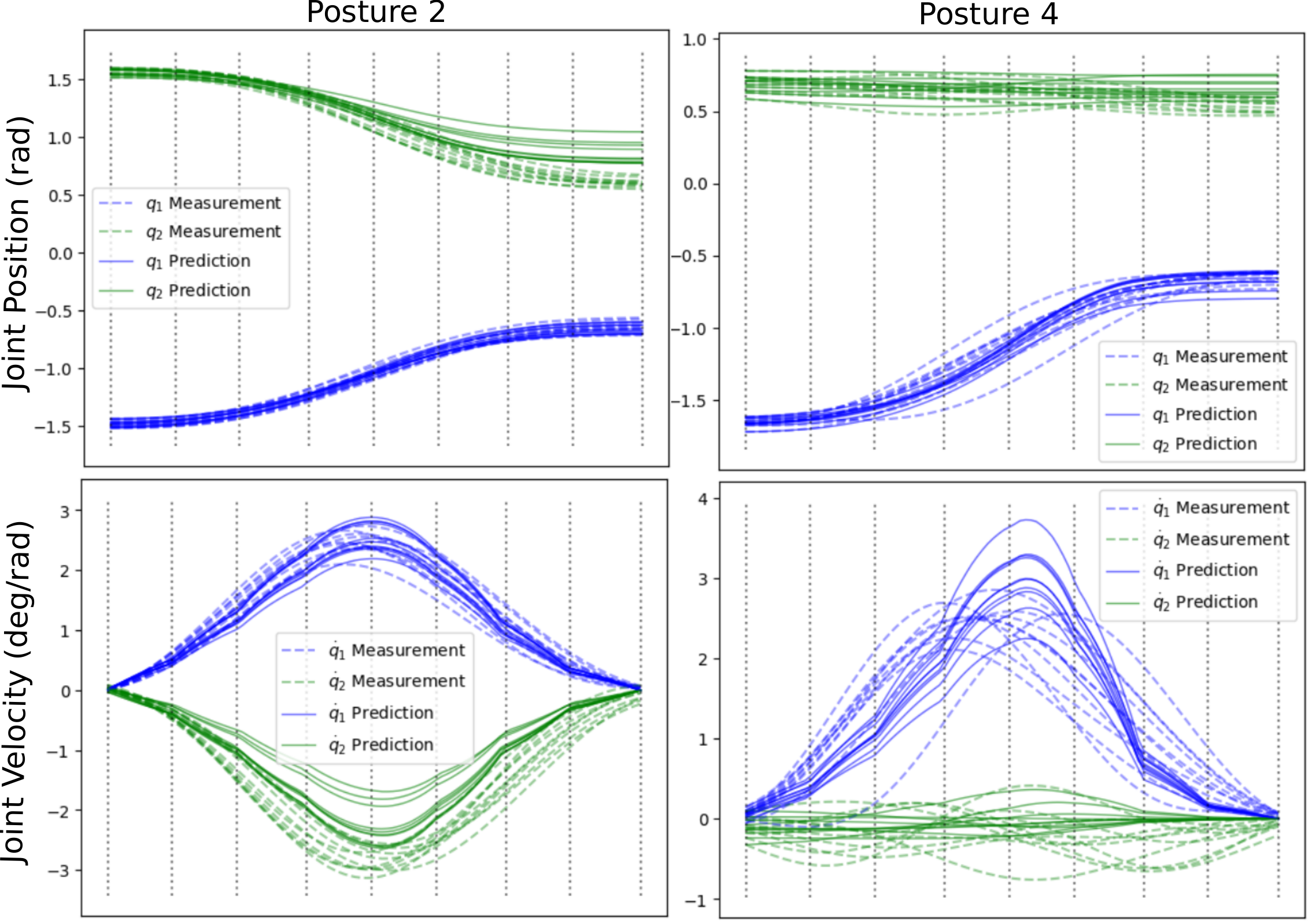}
    \caption{Inter-Subject Cross-Validation of the learned weights (8 sections) by MO-IRL for initial postures 2 and 4. The top row shows the overlayed measured and predicted joint values from the second subject that are not used for the MO-IRL training, where $q_1$ and $q_2$ are shown by blue and green, respectively. Predictions (DOC solutions) and measured trajectories are shown with solid and dashed lines, respectively. The bottom row shows the corresponding joint velocities of the top row trajectories. The trajectories are normalized in length for clearer presentation.}
    \label{fig:is}
\end{figure}

Table~\ref {Table: rmse} shows the corresponding RMSE values for $q_1$, $q_2$, and $\bb{q} = [q_1, q_2]$, after training, cross-validation, and ISCV for 1, 6, and 8-section weights for each initial posture.
When a single section is used, the average RMSE was $10.4$deg, while it was $6.4$deg and $5.6$deg for 6- and 8-sections during training, respectively. 
It indicates that time-varying weights are important although adding more weight sections (from 6 to 8) does not drastically improve the overall prediction average RMSE.

One can see from these results that postures 3 and 5 are more challenging for the MO-IRL to learn, as their RMSE is nearly twice as large as the other postures. This can be attributed to the nature of these initial postures that require the human subject to have more activity in the joint space to result in a fairly small motion in the task space. In other words, in the tasks where the initial elbow angle is more acute than others, more change in the joint space is required to result in a similar task space motion.
Interestingly, both the cross-validation and ISCV trials exhibit average RMSE values that are very similar to those obtained on the training set.
The achieved accuracy, about $8$ degrees for cross validation, is significantly lower than results found in the literature that usually use a single set of weights \cite{berret2011, sylla2014, Lin2021}.

\small
\begin{table*}[b]
    \caption{RMSE on the joint angles (Deg) for 1, 6, and 8 weight sections \\ for training, cross-validation, and ISCV processes}
    
    \begin{center}
        \begin{tabular}{ 
          | >{\raggedright\arraybackslash} p{0.009\textwidth}
          | >{\centering\arraybackslash} p{0.088\textwidth}
          | >{\centering\arraybackslash} p{0.09\textwidth}
          | >{\centering\arraybackslash} p{0.09\textwidth}
          || >{\centering\arraybackslash} p{0.088\textwidth}
          | >{\centering\arraybackslash} p{0.09\textwidth}
          | >{\centering\arraybackslash} p{0.092\textwidth}
          || >{\centering\arraybackslash} p{0.09\textwidth}
          | >{\centering\arraybackslash}p{0.09\textwidth}
          | >{\centering\arraybackslash}p{0.09\textwidth} | }
        \hline
            \hfill & \multicolumn{3}{|c||}{Training (10 trials)} & \multicolumn{3}{|c||}{Cross-Validation (10 trials)} & \multicolumn{3}{|c|}{ISCV (20 trials)} \\ \hline \hline
            \hfill & 1 Section & 6 Sections & 8 Sections & 1 Section & 6 Sections & 8 Sections & 1 Section & 6 Sections & 8 Sections \\ \hline
            
            \hfill & \multicolumn{9}{|c|}{Posture 1} \\ \hline \hline
            $q_1$ & $11.32\pm13.9$ & $5.87\pm8.57$ & $\mathbf{2.54\pm1.23}$ & $8.71\pm3.72$ & $4.15\pm3.76$ & $\mathbf{2.98\pm3.52}$ & $7.99\pm5.91$ & $\mathbf{7.87\pm5.03}$ & $8.94\pm5.04$  \\ \hline
            $q_2$ & $14.65\pm12.4$ & $9.52\pm9.48$ & $\mathbf{5.52\pm4.98}$ & $5.97\pm2.73$ & $5.75\pm5.19$ & $\mathbf{4.99\pm5.29}$ & $9.40\pm2.29$ & $7.94\pm4.24$ & $\mathbf{7.55\pm2.94}$ \\ \hline
            $\mathbf{q}$ & $13.30\pm13.0$ & $8.13\pm8.84$ & $\mathbf{4.41\pm3.48}$ & $7.58\pm2.98$ & $5.23\pm4.28$ & $\mathbf{4.17\pm4.44}$ & $9.13\pm3.60$ & $\mathbf{8.11\pm4.28}$ & $8.50\pm3.63$ \\ \hline
            
            \hfill & \multicolumn{9}{|c|}{Posture 2} \\ \hline \hline
            $q_1$ & $15.10\pm3.61$ & $4.91\pm2.4$ & $\mathbf{4.46\pm2.10}$ & $7.32\pm2.38$ & $\mathbf{3.54\pm2.75}$ & $4.19\pm2.68$ & $9.30\pm4.97$ & $3.58\pm3.80$ & $\mathbf{3.31\pm2.51}$ \\ \hline
            $q_2$ & $13.96\pm2.56$ & $7.16\pm1.74$ & $\mathbf{6.83\pm2.46}$ & $13.88\pm6.96$ & $9.52\pm6.59$ & $\mathbf{8.63\pm2.58}$ & $15.8\pm4.08$ & $\mathbf{8.21\pm3.67}$ & $8.89\pm4.14$ \\ \hline
            $\mathbf{q}$ & $14.55\pm3.08$ & $6.33\pm1.47$ & $\mathbf{5.87\pm2.01}$ & $11.26\pm4.84$ & $7.26\pm4.93$ & $\mathbf{6.89\pm2.34}$ & $13.2\pm3.66$ & $\mathbf{6.56\pm3.31}$ & $6.88\pm3.07$ \\ \hline
            
            \hfill & \multicolumn{9}{|c|}{Posture 3} \\ \hline \hline
            $q_1$ & $11.78\pm3.67$ & $\mathbf{4.53\pm1.65}$ & $7.06\pm3.33$ & $14.31\pm2.60$ & $\mathbf{8.93\pm2.26}$ & $9.85\pm2.55$ & $13.05\pm19.5$ & $\mathbf{8.63\pm4.17}$ & $8.84\pm5.30$ \\ \hline
            $q_2$ & $12.84\pm3.27$ & $8.81\pm2.12$ & $\mathbf{7.21\pm3.78}$ & $17.07\pm3.42$ & $18.64\pm6.68$ & $\mathbf{16.4\pm7.40}$ & $19.52\pm9.57$ & $15.85\pm8.34$ & $\mathbf{15.8\pm10.1}$ \\ \hline
            $\mathbf{q}$ & $12.38\pm3.28$ & $\mathbf{7.12\pm1.24}$ & $7.30\pm3.22$ & $15.87\pm2.32$ & $14.69\pm4.75$ & $\mathbf{13.6\pm5.27}$ & $16.90\pm8.06$ & $\mathbf{13.0\pm6.09}$ & $13.03\pm7.76$ \\ \hline
            
            \hfill & \multicolumn{9}{|c|}{Posture 4} \\ \hline \hline
            $q_1$ & $3.97\pm2.15$ & $3.69\pm1.88$ & $\mathbf{3.47\pm1.85}$ & $2.78\pm1,78$ & $2.95\pm1.70$ & $\mathbf{2.57\pm1.84}$ & $4.67\pm2.14$ & $4.13\pm2.03$ & $\mathbf{4.08\pm2.04}$ \\ \hline
            $q_2$ & $4.12\pm1.94$ & $4.15\pm1.61$ & $\mathbf{3.88\pm1.75}$ & $2.76\pm1.56$ & $3.06\pm1.90$ & $\mathbf{2.53\pm1.48}$ & $4.67\pm2.14$ & $4.03\pm2.84$ & $\mathbf{4.01\pm2.80}$ \\ \hline
            $\mathbf{q}$ & $4.31\pm1.43$ & $4.14\pm1.15$ & $\mathbf{3.83\pm1.47}$ & $3.00\pm1.12$ & $3.27\pm1.28$ & $\mathbf{2.83\pm1.14}$ & $4.57\pm2.02$ & $4.34\pm1.97$ & $\mathbf{4.31\pm1.96}$ \\ \hline
            
            \hfill & \multicolumn{9}{|c|}{Posture 5} \\ \hline \hline
            $q_1$ & $\mathbf{7.05\pm2.84}$ & $8.17\pm2.29$ & $7.83\pm1.62$ & $16.86\pm4.73$ & $9.83\pm5.84$ & $\mathbf{8.26\pm5.67}$ & $13.80\pm7.80$ & $\mathbf{7.86\pm5.33}$ & $9.63\pm6.43$ \\ \hline
            $q_2$ & $9.34\pm4.17$ & $7.54\pm3.71$ & $\mathbf{7.23\pm3.16}$ & $19.12\pm4.95$ & $10.74\pm7.38$ & $\mathbf{9.81\pm5.15}$ & $16.27\pm5.91$ & $\mathbf{10.6\pm4.66}$ & $10.86\pm4.57$ \\ \hline
            $\mathbf{q}$ & $8.38\pm3.32$ & $8.33\pm1.40$ & $\mathbf{7.64\pm2.18}$ & $18.09\pm4.58$ & $10.44\pm6.42$ & $\mathbf{9.33\pm4.96}$ & $15.54\pm5.82$ & $\mathbf{9.80\pm4.11}$ & $10.73\pm4.62$ \\ \hline
            \hfill & \multicolumn{9}{|c|}{ALL Postures} \\ \hline \hline
            $q_1$ & $9.84\pm3.90$ & $5.43\pm1.54$ & $\mathbf{5.07\pm2.05}$ & $10.00\pm5.03$ & $5.88\pm2.90$ & $\mathbf{5.57\pm2.94}$ & $9.76\pm3.36$ & $\mathbf{6.41\pm2.12}$ & $6.96\pm2.69$ \\ \hline
            $q_2$ & $10.98\pm3.89$ & $7.44\pm1.85$ & $\mathbf{6.13\pm1.29}$ & $11.76\pm6.35$ & $9.54\pm5.30$ & $\mathbf{8.47\pm4.73}$ & $13.13\pm5.35$ & $\mathbf{9.33\pm3.88}$ & $9.42\pm3.89$ \\ \hline
            $\mathbf{q}$ & $10.41\pm3.81$ & $6.43\pm1.42$ & $\mathbf{5.60\pm1.57}$ & $10.88\pm5.51$ & $7.71\pm3.86$ & $\mathbf{7.02\pm3.77}$ & $11.45\pm4.27$ & $\mathbf{7.87\pm2.80}$ & $8.19\pm2.93$ \\ \hline
        \end{tabular}
    \label{Table: rmse}
    \end{center}
\end{table*}
\normalsize

Fig.~\ref{fig:weights} shows the learned weights for various postures and sections (the weights are normalized for better presentation). The varying weights in both 6- and 8-section cases, supported by the high RMSE observed with single-section weights, corroborate the fact that one uniform weight set might not be enough to correctly explain the intent of humans, even in a simple task like reaching a wall. 
In contrast to Berret et al. \cite{berret2011}, the energy-related cost function was only marginally observed in postures 3, 4, and 5 of the investigated trials. This discrepancy may stem from the differing normalization techniques employed. While Berret et al. used the so-called pivot method \cite{panchea2018}, our proposed approach does not enforce normalization, thereby allowing greater flexibility in the inferred cost weights.

Nevertheless, as shown in Fig. \ref{fig:weights}, joint acceleration minimization appears to play a dominant role, especially during the initiation and termination phases of the movement, suggesting a strategy aimed at ensuring smooth motion onset and precise stopping. The movement onset likely reflects a strategy to ensure a smooth and stable initiation, avoiding abrupt or energetically costly changes in motor commands. This result is also consistent with prior studies showing that reaching trajectories typically exhibit bell-shaped velocity profiles and low jerk \cite{flash1985}, suggesting implicit optimization of higher-order derivatives of position.

Reducing acceleration towards the end of the motion may serve to finely tune the final position, increasing precision, and ensuring comfortable deceleration before reaching the target. These findings support the notion that motor control is not governed by a single, static cost function across the entire trajectory, but rather by a dynamic trade-off between competing criteria such as effort minimization, accuracy, and smoothness, adapted to the temporal structure of the task. 

Fig. \ref{fig:is} shows the ISCV results for postures 2 and 4 by comparing the actual human joint position and velocity trajectories of the second subject, with the predicted trajectories solved by DOC using the learned weights. This figure shows that not only do the predictions look very similar to the measured human motion, but also the scatteredness of the predicted trajectories resemble that of the measured trajectories. This preliminary test suggests that the learned behavior from one subject is transposable to other individuals, while retaining the variability expected from human behavior. 

Another notable observation is that while human trajectories (especially in joint velocity) differ from one trial to another, even for a single subject and posture, the DOC predictions are deterministic and only depend on initial conditions $\bb{x}(0)$. The weights learned capture an average behavior based on the 10 trials it was trained on. We posit that this helps learn key common feature from the movements while discarding less relevant variations. We can see this effect in Fig. \ref{fig:is} where the predictions do not necessarily match joint velocity profiles, but they all reach for the goal, same as the human intended to.


To the best of our knowledge, this study is the first to perform actual ISCV. While conclusions should be drawn with caution, as only one subject was included in the ISCV test, the results are promising: the obtained RMSE in ISCV is of the same order of magnitude as in the training and cross-validation trials. This may indicate that the identified weights generalize well and are not overfitted despite the use of time-varying weights. This is likely achieved thanks to the use of joint velocity directly in the MO-IRL's step search and also the use of weight regularization ($L2$ norm in Eq.\eqref{'moirl'}).


%% file: conclusion/conclusion.tex
This paper proposes a framework based on MO-IRL to predict human arm-reaching motions using time-varying weights. Our empirical results demonstrate the benefits of using time-varying weights in the cost function to learn human movements with explainable cost functions.  Importantly, we achieved lower reconstruction errors than previously reported in the literature. Our results suggest a strong emphasis on joint acceleration at the beginning and end of the movement while other cost features appear less dominant. This is consistent with the idea that humans will generally refrain from highly accelerated movements in the initial and terminal stages of the task. We also provided preliminary results for inter-subject analysis of the resulting cost function, which exhibits promising results on the ability to generalize the learned cost function from one subject to another. The reported RMSE values for the cross-validation and their close similarities to the ISCV results show promises towards a generalizable IRL framework for understanding human task intentions and also towards transferring human movements onto humanoids. 

Our future work includes:
\begin{itemize}
    \item Testing the algorithm on a higher number of subjects to analyze its generalizability. 
    \item Conducting sensitivity and occlusion tests to further understand the effect of individual cost features.
    \item Incorporating multimodal step-acceptance in MO-IRL (based on exerted force, joint torque, etc.) to improve convergence.
    \item Testing the framework on more complex tasks that include more dynamic behaviors. 
\end{itemize}